\documentclass[10pt,twocolumn,letterpaper]{article}

\usepackage{cvpr}              

\usepackage[dvipsnames]{xcolor}

\definecolor{cvprblue}{rgb}{0.21,0.49,0.74}
\usepackage[pagebackref,breaklinks,colorlinks,citecolor=cvprblue]{hyperref}
\usepackage{multirow,booktabs,pifont}
\usepackage{amsmath} 
\DeclareMathOperator*{\argmax}{arg\,max}
\def\paperID{*****} 
\def\confName{CVPR}
\def\confYear{2024}

\title{ALGO: Object-Grounded Visual Commonsense Reasoning for Open-World Egocentric Action Recognition}

\author{Sanjoy Kundu\\
Auburn University\\
Auburn, Alabama, USA\\
{\tt\small szk0266@auburn.edu}
\and
Shubham Trehan\\
Auburn University\\
Auburn, Alabama, USA\\
{\tt\small szt0113@auburn.edu}
\and
Sathyanarayanan N Aakur\\
Auburn University\\
Auburn, Alabama, USA\\
{\tt\small san0028@auburn.edu}
}

\begin{document}
\maketitle
\begin{abstract}
Learning to infer labels in an open world, i.e., in an environment where the target ``labels'' are unknown, is an important characteristic for achieving autonomy. Foundation models pre-trained on enormous amounts of data have shown remarkable generalization skills through prompting, particularly in zero-shot inference. 
However, their performance is restricted to the correctness of the target label's search space. 
In an open world, this target search space can be unknown or exceptionally large, which severely restricts the performance of such models. To tackle this challenging problem, we propose a neuro-symbolic framework called ALGO - \underline{A}ction \underline{L}earning with \underline{G}rounded \underline{O}bject recognition that uses symbolic knowledge stored in large-scale knowledge bases to infer activities in egocentric videos with limited supervision using two steps. First, we propose a neuro-symbolic prompting approach that uses \textit{object-centric} vision-language models as a noisy oracle to ground objects in the video through evidence-based reasoning. Second, driven by prior commonsense knowledge, we discover plausible activities through an energy-based symbolic pattern theory framework and learn to ground knowledge-based action (verb) concepts in the video. Extensive experiments on four publicly available datasets (EPIC-Kitchens, GTEA Gaze, GTEA Gaze Plus) demonstrate its performance on open-world activity inference.
\end{abstract}    
\section{Introduction}
\label{sec:intro}
Humans display a remarkable ability to recognize unseen concepts (actions, objects, etc.) by associating known concepts gained through prior experience and reasoning over their attributes. Key to this ability is the notion of ``grounded'' reasoning, where abstract concepts can be mapped to the perceived sensory signals to provide evidence to confirm or reject hypotheses. In this work, we aim to create a computational framework that tackles open-world egocentric activity understanding. We define an activity as a complex structure whose semantics are expressed by a combination of actions (verbs) and objects (nouns). To recognize an activity, one must be cognizant of the object label, action label, and the possibility of any combination since not all actions are plausible for an object.
Supervised learning approaches~\cite{wang2013action,sigurdsson2018actor,ma2016going,dosovitskiyimage} have been the dominant approach to activity understanding but are trained in a ``closed'' world, where there is an implicit assumption about the target labels. The videos during inference will always belong to the label space seen during training.
Zero-shot learning approaches~\cite{zhang2017first,lin2022egocentric,zhao2023lavila,ashutosh2023hiervl} relax this assumption by considering disjoint ``seen'' and ``unseen'' label spaces where all labels are not necessarily represented in the training data. This setup is a \textit{known} world, where the target labels are pre-defined and aware during training.
In this work, we define an \textit{open} world to be one where the target labels are unknown during both training and inference. The goal is to recognize elementary concepts and infer the activity.
\begin{figure*}[t]
    \centering
    \includegraphics[width=0.9\textwidth]{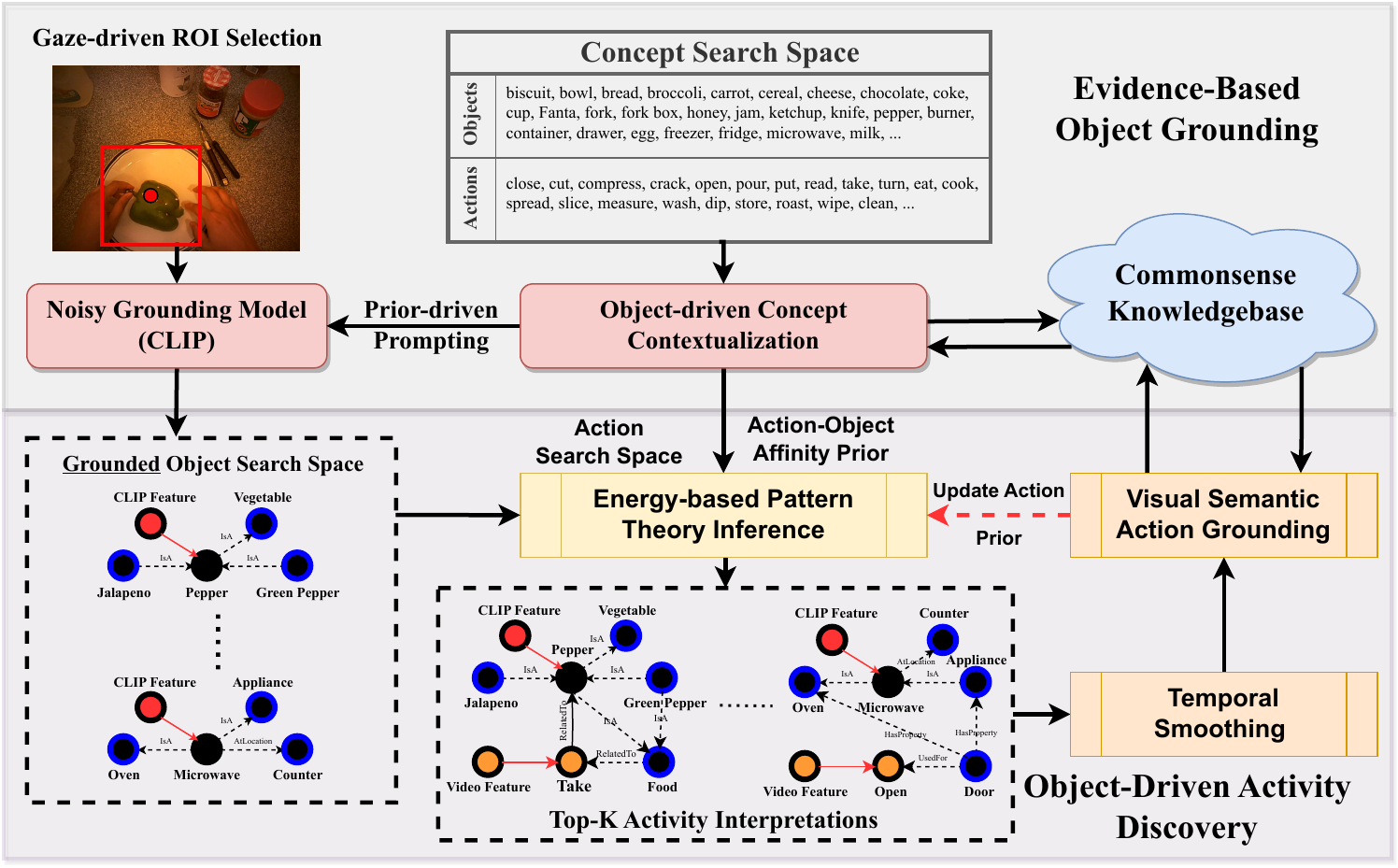}
    \caption{\textbf{Overall architecture} of the proposed approach (ALGO) is illustrated here. Using a two-step process, we first \textit{ground} the objects within a gaze-driven ROI using CLIP~\cite{radford2021learning} as a noisy oracle before reasoning over the plausible activities performed in the video. }
    \label{fig:arch}
\end{figure*}
%
We propose to tackle this problem using a neuro-symbolic framework that leverages advances in multi-modal foundation models to ground concepts from symbolic knowledge bases, such as ConceptNet~\cite{speer2017conceptnet}, in visual data. The overall approach is shown in Figure~\ref{fig:arch}. Using the energy-based pattern theory formalism~\cite{aakur2022knowledge,aakur2019generating,grenander1996elements} to represent symbolic knowledge, we ground objects (nouns) using CLIP~\cite{radford2021learning} as a noisy oracle. Driven by prior knowledge, novel activities (verb+noun) are inferred, and the associated action (verb) is grounded in the video to learn visual-semantic associations for novel, unseen actions.

The \textbf{contributions} of this work are three-fold: (i) We present a neuro-symbolic framework to leverage compositional properties of objects to prompt CLIP for evidence-based grounding. (ii) We propose object-driven activity discovery as a mechanism to reason over prior knowledge and provide action-object affinities to constrain the search space. (iii) We demonstrate that the inferred activities can be used to ground unseen actions (verbs) from symbolic knowledge in egocentric videos, which can generalize to unseen and unknown action spaces.

\textbf{Egocentric video analysis} has been extensively explored in computer vision literature, having applications in virtual reality~\cite{han2020megatrack} and human-machine interaction.
While Supervised learning has been the dominant approach such as ~\cite{Sudhakaran_2019_CVPR},~\cite{ma2016going},~\cite{Zhou_2016_CVPR,Wang_2021_ICCV},~\cite{Ryoo_2015_CVPR} along with some zero-shot learning approaches ~\cite{zhang2017first,sigurdsson2018actor}, 
KGL~\cite{aakur2022knowledge} is one of the first works to address the problem of \textbf{open-world understanding}. They represent knowledge elements derived from ConceptNet~\cite{speer2017conceptnet}, using pattern theory~\cite{aakur2019generating,desouza2016pattern,grenander1996elements}. Their method depends on an object detector to link objects in a source domain before translating concepts to the target space via ConceptNet-based semantic connections. However, this approach has drawbacks: (i) false alarms may arise if the initial object detector misses the object, resorting to the \textit{closest} object instead, and (ii) it relies on ConceptNet for correspondences, potentially disregarding objects with zero corresponding probabilities.
The development of object-centric foundation models has enabled impressive capabilities in zero-shot object recognition in images, as demonstrated by CLIP~\cite{radford2021learning}, DeCLIP~\cite{li2022supervision}, and ALIGN~\cite{jia2021scaling}. 
Recent works, such as EGO-VLP~\cite{lin2022egocentric}, Hier-VL~\cite{ashutosh2023hiervl}, LAVILLA~\cite{zhao2023lavila}, and CoCa~\cite{yu2022coca} have expanded the scope of multimodal foundation models to include egocentric videos and have achieved impressive performance in zero-shot generalization which requires substantial amounts of curated pre-training data to learn semantic associations among concepts.
\textbf{Neuro-symbolic models}~\cite{nye2021improving,NEURIPS2020_94c28dcf,NEURIPS2022_3ff48dde,aakur2022knowledge} show promise in reducing the increasing dependency on data. We extend the idea of neuro-symbolic reasoning to address egocentric, open-world activity recognition.
\section{Proposed Framework: ALGO}
\textbf{Problem Formulation.}Our task is to recognize unknown activities in egocentric videos within an open-world setting. We aim to develop a framework that identifies elementary concepts, establishes semantic associations, and effectively combines these to interpret the observed activity. Activities are formed by combining concepts from two distinct sets: an object (nouns) ($G_{obj}$) and an action (verbs) ($G_{act}$) drawn from a predefined search space.

\textbf{Overview.} Our proposed framework ALGO (Action Learning with Grounded Object recognition), as illustrated in Figure~\ref{fig:arch} tackles the problem of discovering novel actions in an open world. 
It starts by hypothesizing the plausible objects through evidence-based object grounding (Section~\ref{sec:NSG}) by exploring prior knowledge from a symbolic knowledge base. An energy-based inference mechanism (Section~\ref{sec:inference}) then identifies the plausible actions on these objects. We leverage visual-semantic action grounding to discover activities without explicit supervision by employing tools like CLIP~\cite{radford2021learning} and ConceptNet~\cite{speer2017conceptnet}, respectively. 

\textbf{Knowledge Representation.} We use Grenander's pattern theory ~\cite{grenander1996elements} to represent the knowledge, integrating neural and symbolic elements in a unified, energy-based representation. 
We refer the reader to Aakur \textit{et al.}~\cite{aakur2019generating} and de Souza \textit{et al.} ~\cite{desouza2016pattern} for a deeper exploration of knowledge representation in pattern theory.

\subsection{Evidence-based Object Grounding}\label{sec:NSG}
The first step to assess the plausibility of object concepts (generators $\{g^o_1, g^o_2, \ldots g^o_i\} \in G_{obj}$) by \textit{grounding} them in the input video \textit{$V_i$}. \textit{Grounding} gathers evidence to support or reject a concept's presence. 
 To enhance object recognition accuracy, we propose a neuro-symbolic mechanism that leverages compositional properties from ConceptNet to compute the likelihood of an object's presence. This involves constructing an ego-graph for each object and using CLIP to evaluate the likelihood of ungrounded generators, using prior knowledge to assess the presence of grounded object generators. 

Given this set of \textit{ungrounded} generators ($\{\bar{g}^o_i\} \forall g^o_i \in G_{obj}$), we then prompt CLIP to provide likelihoods for each ungrounded generator $p(\bar{g}^o_i \vert I_t)$ to compute the \textit{evidence-based} likelihood for each \textit{grounded} object generator $\underline{g}^o_i$ as defined by the probability \( p(\underline{g}^o_i \vert \bar{g}^o_i, I_t, K_{CS}) = p(\underline{g}^o_i | I_t)*\left\lVert \sum_{\forall \bar{g}^o_i} p({g}^o_i, \bar{g}^o_i \vert E_{{g}^o_i}) * p(\bar{g}^o_i) | I_t) \right\rVert^2 \), where $p({g}^o_i, \bar{g}^o_i \vert E_{{g}^o_i})$ is the edge weight from the edge graph $E_{{g}^o_i}$ (sampled from a knowledge graph $K_{CS}$) that acts as a prior for each ungrounded evidence generator $\bar{g}^o_i$, and $p(\bar{g}^o_i) | I_t)$ is the likelihood from CLIP for its presence in each frame $I_t$.
%
To focus on relevant objects, we use the human gaze to select a specific region for analysis, leveraging object-grounding insights.

\subsection{Object-driven Activity Discovery}\label{sec:inference}
The next step focuses on identifying plausible activities in the video by considering object affordances and the compatibility of action-object pairs using prior knowledge.
The probability of an activity (defined by an action generator \( g^a_i \) and a grounded object generator \( \underline{g}^o_j \)) is given by \( p(g^a_i, \underline{g}^o_j \vert K_{CS}) = \argmax_{\forall E \in K_{CS}} \sum_{(\bar{g}_{m}, \bar{g}_{n}) \in E}{w_k * K_{CS}(\bar{g}_{m}, \bar{g}_{n})} \). where $E$ is the collection of all paths between $g^a_i$ and $\underline{g}^o_j$ in a commonsense knowledge graph $K_{CS}$, $w_i$ is a weight drawn from an exponential decay function based on the distance of the node $\bar{g}_n$ from $g^a_i$. After filtering for compositional properties, the path with the maximum weight is chosen with the optimal action-object affinity.

\textbf{Energy-based Activity Inference.} 
To infer activities, we assign an energy term to each label using configurations composed of generators connected by affinity-based bonds. Each configuration includes a grounded object generator ($\underline{g}^o_i$), its ungrounded evidence generators ($\bar{g}^o_j$), an action generator ($g^a_k$), and related ungrounded generators, structured by a graph derived from ConceptNet. The energy of a configuration $c_i$ is expressed as:
\begin{equation}
\begin{split}
    E(c) &= \phi(p(\underline{g}^o_i \vert \bar{g}^o_j, I_t, K_{CS})) + \phi(p(g^a_k, \underline{g}^o_i \vert K_{CS})) \\ 
    &+ \phi(p(g^a_k | I_t))
\end{split}
    \label{eqn:configEnergy}
\end{equation}
Hence, activity inference becomes an optimization over Equation~\ref{eqn:configEnergy} to find the configuration (or activity interpretation) with the least energy.
\subsection{Visual-Semantic Action Grounding}\label{sec:temporalGround}
In this step we aim to map inferred action verbs into a semantic embedding space provided by ConceptNet Numberbatch, using a linear projection to translate visual features from the video to 300-dimensional semantic vectors ($\mathbb{R}^{1\times300}$). This process involves training a mapping function $\psi(g^a_i, f_V)$, primarily using a mean squared error (MSE) loss, to ground actions recognized in the video within the broader semantic context of ConceptNet.

\begin{table*}[t]
    \centering
    \begin{tabular}{|c|c|c|c|c|c|c|c|c|}
    \toprule
    \multirow{2}{*}{\textbf{Approach}}& \multirow{2}{*}{\textbf{Search }} & \multirow{2}{*}{\textbf{VLM?}} & \multicolumn{3}{|c|}{\textbf{GTEA Gaze}} & \multicolumn{3}{|c|}{\textbf{GTEA GazePlus}}\\
    \cmidrule{4-9}
    & \textbf{Space} &  & \textbf{Object} & \textbf{Action} & \textbf{Activity} & \textbf{Object} & \textbf{Action} & \textbf{Activity}\\
    \toprule
    Two-Stream CNN~\cite{simonyan2014two} & Closed & \ding{55} & 38.05 & 59.54 & \underline{53.08} & \underline{61.87} & 58.65 & 44.89\\
    IDT~\cite{wang2013action} & Closed & \ding{55} & \underline{45.07} & \underline{75.55} & 40.41 & 53.45 & \underline{66.74} & \underline{51.26}\\
    Action Decomposition~\cite{zhang2017first} & Closed & \ding{55} & \textbf{60.01} & \textbf{79.39} & \textbf{55.67} & \textbf{65.62} & \textbf{75.07} & \textbf{57.79}\\
    \midrule
    Random & Known & \ding{55} & 3.22 & 7.69 & 2.50 & 3.70 & 4.55 & 2.28\\
    Action Decomposition ZSL~\cite{zhang2017first} & Known & \ding{55} & \underline{40.65} & \textbf{85.28} & \textbf{39.63} & \underline{43.44} & \underline{27.68} & \underline{15.98}\\
    ALGO ZSL (Ours) & Known & \ding{55} & \textbf{49.47} & \underline{74.74} & \underline{27.34} & \textbf{47.67} & \textbf{29.31} & \textbf{16.68}\\
    \midrule
    KGL~\cite{aakur2022knowledge} & Open & \ding{55} &  5.12 & 8.04 & 4.91 & 14.78 & 6.73 & 10.87\\
    KGL+CLIP~\cite{aakur2022knowledge} & Open & \ding{55} &  {10.36} & {8.15} & {9.21} & {20.49} & {9.23} & {14.86}\\
    ALGO (Ours) & Open & \ding{55} &  \textbf{13.07} & \textbf{17.05} & \textbf{15.05} & \textbf{26.23} & \textbf{11.44} & \textbf{18.84}\\
    \midrule
    EgoVLP~\cite{lin2022egocentric} & Open & \ding{51} &  10.17 & 8.45 & 9.31 & 29.43 & 17.17 & 23.30 \\
    LaViLa~\cite{zhao2023lavila} & Open & \ding{51} &  6.07 & 23.07 & 14.57 & 28.27 & 25.47 & 26.87 \\
    ALGO+EgoVLP & Open & \ding{51} &  {8.61} & {4.64} & {6.63} & {20.48} & {20.48} & {20.48}\\
    ALGO+LaViLa & Open & \ding{51} &  \textbf{17.50} & \textbf{26.60} & \textbf{22.05} & \textbf{30.74} & \textbf{27.00} & \textbf{28.87}\\
    \bottomrule
    \end{tabular}
    \caption{\textbf{Open-world activity recognition} performance on the GTEA Gaze and GTEA Gaze Plus datasets. We compare approaches with a closed search space, those with a known search space, and those with a partially open one. Accuracy is reported for predicted objects, actions, and activities. VLM: Vision-Language Model pre-trained on egocentric video data. * indicates training on ``seen'' classes from the same dataset(s) and leave-one-action-out evaluation.}
    \label{tab:gaze_results}
    \vspace{-5mm}
\end{table*}

\textbf{Temporal Smoothing} 
We implement temporal smoothing by first aggregating action predictions at the frame level. For each frame, we compute the top five actions based on their energy levels, then average these across the clip to stabilize the learning process. This aggregated data forms the basis for training the mapping function $\psi(g^a_i, f_V)$, focusing on the most frequent and energetically consistent actions.


\textbf{\textbf{Posterior-based Activity Refinement.}}\label{sec:posterior}
The final step involves an iterative refinement process that updates the action concept priors based on predictions from the visual-semantic grounding mechanism (Section~\ref{sec:temporalGround}). We adjust the action priors in the energy computation (Equation~\ref{eqn:configEnergy}), re-ranking activity interpretations to reflect clip-level dynamics better. The refinement cycle alternates between updating posterior probabilities and re-training the action grounding model until generalization error saturates.

\section{Experimental Evaluation}\label{sec:results}

\textbf{Data.} We evaluate the approach on GTEA Gaze~\cite{fathi2012learning}, GTEA GazePlus~\cite{li2013learning}, and EPIC-Kitchens-100~\cite{Damen2021PAMI,Damen2022RESCALING} datasets, which contain egocentric, multi-subject videos of meal preparation activities. The GTEA Gaze dataset has 10 verbs and 38 nouns (search space of 380 activities), while GTEA GazePlus has 15 verbs and 27 nouns (search space of 405), Charades-Ego has 33 verbs and 38 nouns (search space of 1254), and Epic-Kitchens has 97 verbs and 300 nouns (search space of 29100). 


\textbf{Baselines.} We compare against both closed-world learning and open-world setup (KGL)~\cite{aakur2022knowledge}. We also create a baseline called ``KGL+CLIP'' by augmenting KGL with CLIP-based grounding by including CLIP's similarity score for establishing semantic correspondences. We compare with supervised learning models as well as zero-shot versions of Action Decomposition and large vision-language models, such as EGO-VLP~\cite{lin2022egocentric}, HierVL~\cite{ashutosh2023hiervl}, and LAVILA~\cite{zhao2023lavila} in both zero-shot and open-world settings.

\begin{table}[t]
    \centering
    \begin{tabular}{|c|c|c|c|c|}
    \toprule
         \textbf{Approach} & \textbf{VLM?} & \textbf{Action} & \textbf{Object} & \textbf{Activity}\\
    \toprule
    Random & \ding{55} & 1.03 & 0.33 & 0.68\\
    KGL~\cite{aakur2022knowledge} & \ding{55} & 3.89 & 2.56 & 3.23\\
    KGL+CLIP~\cite{aakur2022knowledge} & \ding{55} & \underline{5.32} & \underline{4.67} & \underline{4.99}\\
    ALGO (Ours) & \ding{55} & \textbf{10.21} & \textbf{6.76} & \textbf{8.48}\\
    \midrule
    EgoVLP~\cite{lin2022egocentric} & \ding{51} & 10.77 & 19.51 & 15.14\\
    LaViLa~\cite{zhao2023lavila} & \ding{51} & 11.16 & \textbf{23.25} & \underline{17.21}\\
    ALGO+LaViLa & \ding{51} & \textbf{12.54} & \underline{22.84} & \textbf{17.69}\\
    \bottomrule
    \end{tabular}
    \caption{Evaluation on the EPIC-Kitchens-100 dataset. VLM: Vision-Language pre-training on egocentric data. Accuracy for actions, objects, and activity are reported.}
    \label{tab:ek100}
    \vspace{-5mm}
\end{table}
\subsection{Open World Activity Recognition}\label{sec:openWorld}
Table~\ref{tab:gaze_results} summarizes the evaluation results under the open-world inference setting. Top-1 prediction results are reported for all approaches. As can be seen, CLIP-based grounding significantly improves the performance of object recognition for KGL, as opposed to the originally proposed, prior-only correspondence function. However, our neuro-symbolic grounding mechanism (Section~\ref{sec:NSG}) improves it further, achieving an object recognition performance of $13.07\%$ on Gaze and $26.23\%$ on Gaze Plus. 
Similarly, the posterior-based action refinement module (Section~\ref{sec:posterior}) helps achieve a top-1 action recognition performance of $17.05\%$ on Gaze and $11.44\%$ on Gaze Plus, outperforming KGL ($8.04\%$ and $6.73\%$). Adding action priors from LaViLa ($\phi(p(g^a_k | I_t))$ in Equation~\ref{eqn:configEnergy}) allows us to improve the performance further, as indicated by ALGO+LaViLa. 

We also evaluate our approach on the Epic-Kitchens-100 dataset, a larger-scale dataset with a significantly higher number of concepts (actions, verbs, and activities). Table~\ref{tab:ek100} summarizes the results. We significantly outperform non-VLM models while offering competitive performance to the VLM-based models. We see that even without \textit{any video-based training data}, we achieve an action accuracy of $10.21\%$ and object accuracy of $6.76\%$, indicating that we can learn affordance-based relationships for discovering and grounding novel actions in egocentric data. 
\section{Discussion, Limitations, and Future Work}
In this work, we proposed ALGO, a neuro-symbolic framework for open-world egocentric activity recognition that aims to learn novel action and activity classes without explicit supervision. While showing competitive performance, there are two key limitations: (i) it is restricted to ego-centric videos due to the need to navigate clutter by using human attention as a contextual cue for object grounding, and (ii) it requires a knowledge base such as ConceptNet to learn associations between actions and objects.
In the future, we aim to explore attention-based mechanisms~\cite{mounir2023towards,aakur2022actor} to extend the framework to third-person videos and using abductive reasoning~\cite{aakur2023leveraging,Zellers_2019_CVPR} with neural knowledgebase completion models~\cite{bosselut2019comet} to integrate visual commonsense into the reasoning.

\textbf{Acknowledgements.} This research was supported in part by the US National Science Foundation grants IIS 2348689, and IIS 2348690. We thank Dr. Anuj Srivastava (FSU) and Dr. Sudeep Sarkar (USF) for their thoughtful feedback during the discussion about the project's problem formulation and experimental analysis phase. 

{
    \small
    \bibliographystyle{ieeenat_fullname}
    \bibliography{main}
}


\end{document}